\documentclass[sigconf]{acmart}

\usepackage{booktabs} 
\usepackage{enumitem}
\usepackage{subcaption}

\usepackage[ruled]{algorithm2e} 
\usepackage{float}

\SetAlFnt{\small}
\SetAlCapFnt{\small}
\SetAlCapNameFnt{\small}
\SetAlCapHSkip{0pt}

\newcommand{\shortsectionref}[1]{\hyperref[#1]{\S\ref*{#1}}}

\copyrightyear{2022}
\acmYear{2022}
\setcopyright{acmlicensed}
\acmConference[SIGGRAPH '22 Conference Proceedings]{Special Interest Group on Computer Graphics and Interactive Techniques Conference Proceedings}{August 7--11, 2022}{Vancouver, BC, Canada}
\acmBooktitle{Special Interest Group on Computer Graphics and Interactive Techniques Conference Proceedings (SIGGRAPH '22 Conference Proceedings), August 7--11, 2022, Vancouver, BC, Canada}
\acmPrice{15.00}
\acmDOI{10.1145/3528233.3530728}
\acmISBN{978-1-4503-9337-9/22/08}

\begin{document}
\title{Learning to Brachiate via Simplified Model Imitation}

\author{Daniele Reda}
\orcid{0000-0002-7101-0519}
\email{dreda@cs.ubc.ca}
\affiliation{
 \institution{University of British Columbia}
 \city{Vancouver}
 \country{Canada}
}
\authornote{Indicates equal contribution.}

\author{Hung Yu Ling}
\email{hyuling@cs.ubc.ca}
\affiliation{
 \institution{University of British Columbia}
 \city{Vancouver}
 \country{Canada}
}
\authornotemark[1]

\author{Michiel van de Panne}
\email{van@cs.ubc.ca}
\affiliation{
 \institution{University of British Columbia}
 \city{Vancouver}
 \country{Canada}
}

\begin{abstract}
Brachiation is the primary form of locomotion for gibbons and siamangs, in which these primates swing from tree limb to tree limb using only their arms. 
It is challenging to control because of the limited control authority, the required advance planning,
and the precision of the required grasps.
We present a novel approach to this problem using reinforcement learning, 
and as demonstrated on a finger-less 14-link planar model that learns to brachiate across challenging handhold sequences.
Key to our method is the use of a simplified model, a point mass with a virtual arm,
for which we first learn a policy that can brachiate across handhold sequences with a prescribed order.
This facilitates the learning of the policy for the full model, for which it provides guidance
by providing an overall center-of-mass trajectory to imitate, as well as for the timing of the holds.
Lastly, the simplified model can also readily be used for planning suitable sequences of handholds 
in a given environment.
Our results demonstrate brachiation motions with a variety of durations for the flight and hold phases,
as well as emergent extra back-and-forth swings when this proves useful.
The system is evaluated with a variety of ablations.
The method enables future work towards more general 3D brachiation,
as well as using simplified model imitation in other settings.
For videos, supplementary material and code, visit:
\\\url{https://brachiation-rl.github.io/brachiation}.
\end{abstract}

%
%


%
%


\begin{teaserfigure}
  \centering
  \includegraphics[width=.95\textwidth]{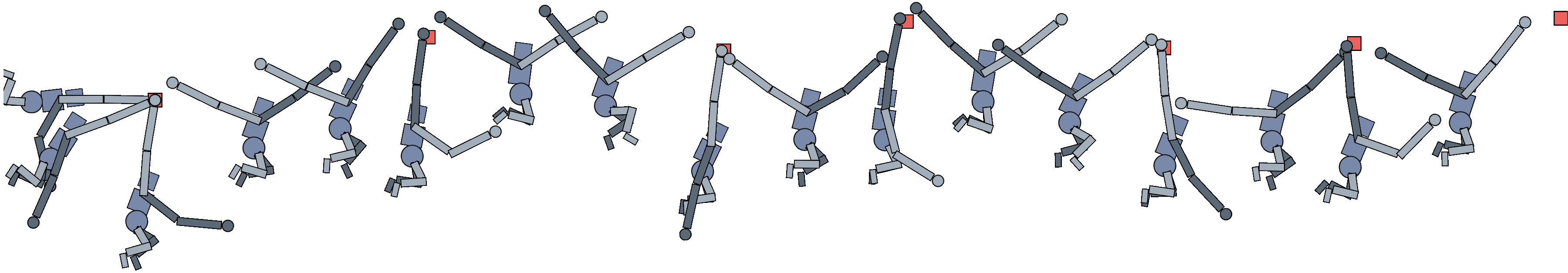}
  \caption{We use a two-stage simplified model imitation learning approach to produce realistic physics-based brachiation motions. The learned control policies can traverse challenging handhold sequences and can demonstrate emergent \textit{pumping} behavior to build momentum for large gaps.}
  \label{fig:teaser}
\end{teaserfigure}
\maketitle

\section{Introduction}

Brachiation is the form of locomotion that consists of moving between tree branches or handholds using the arms 
alone~\cite{fleagle_dynamics_1974, preuschoft_influence_1985}.
Gibbons and siamangs make this seem effortless and are among the world's most agile brachiators.
Brachiating movements can be classified into two types: {\em continuous contact brachiation}, characterized by the animal mantaining contact at all time with a handhold such as a tree branch, and 
{\em ricochetal brachiation}, which involves a flight phase between  
successive grasp~\cite{bertram_point-mass_1999, wan_non-horizontal_2015}. 
These two modes of brachiation are analogous to walking and running, respectively, with ricochetal brachiation used at faster speeds.

Machine learning techniques such as deep reinforcement learning (RL) have been previously applied to legged, aerial, and underwater locomotion~\cite{Aerobatics_Won, CARL_Luo, Min:2019:SoftCon, 2020-SCA-allsteps}.
Despite the apparent similarity to these types of locomotion, brachiation is unique in a number of ways.  
First, the choice of handholds is often discrete, i.e., the choice of particular branches, and therefore unlike the continuous terrain usually available for legged locomotion.
This offers apparently fewer features to control, i.e., less 
control authority, while at the same time necessitating high spatial precision to effect a grasp.
Second, advance planning of the motion across a sequence of handholds becomes more important because of the momentum that may be needed to reach future handholds. 
Lastly, brachiation offers the possibility of very efficient horizontal motion;  an ideal
point mass can follow alternating pendulum-like swings and parabolic flight phases with no loss
of energy~\cite{bertram_point-mass_1999}.

In our work, we present a fully learned solution for simulating a 14-link planar articulated model.
Similar to previous work on brachiation, our model uses a pinned attachment model as a proxy for the grasping mechanics.
The learned control policy demonstrates ricochetal brachiation across challenging sequences of handholds that, to the best of our knowledge, is the most capable of its kind.

\noindentparagraph{\normalfont Our primary contributions are as follows:}
\begin{itemize}[leftmargin=\parindent]
    \setlength\itemsep{3pt}
    \item We present an effective learning-based solution for planar brachiation.
    The learned control policies can traverse challenging sequences of handholds using ricochetal motions with
    significant flight phases, and demonstrate emergent phenomena, such as extra back-and-forth swings when needed.
    
    \item We propose a general two-stage RL method for leveraging simplified physical models.
    RL on the simplified model allows for fast and efficient exploration, quickly producing an effective control policy.
    The full model then uses imitation learning of the motions resulting from
    the simplified-model for efficient learning of complex behaviors.
\end{itemize}

\section{Related Work}

Our work sits at the intersection of physics-based character animation, 
the study of brachiation, and a broad set of methods that leverage abstract models
in support of motion planning and control.

\subsection{Brachiation}

Due to the fascinating and seemingly effortless nature of brachiation, 
there exists a long history of attempts to reproduce this often-graceful type of motion,
both in simulation and on physical robots. 
The key challenge lies in developing control strategies that are well suited to the underactuated 
and dynamic nature of the task.

A starting point is to first study brachiation in nature, such as via a detailed film study~\cite{fleagle_dynamics_1974}. 
Among other insights, they note that the siamang is able to limit lateral motion of the center of mass
between handholds, in part due to extensive rotation at the wrist, elbow, and shoulder.
A benefit of longer arms for reduced energetic costs of brachiation can also be found~\cite{preuschoft_influence_1985}.
A more recent study~\cite{michilsens_functional_2009} notes that ``compared with other primates, the elbow flexors of gibbons are particularly powerful,
suggesting that these muscles are particularly important for a brachiating lifestyle.''
We note that the limited lateral motion and strong elbow flexors correspond well to the capabilities afforded by sagittal-plane point-mass models.

One of the earliest results for brachiation capable of varying height and distance
uses an intricate heuristic procedure and manually structured motion phases, as applied
to a two-link robot with no flight phases~\cite{saito1992movement,saito_swing_1994}. 
A loosely comparable approach is proposed for a 3-link planar model, restricted to horizontal motions~\cite{zhang_details_1999}.
A point mass model can be shown to produce solutions with zero energetic cost
for both continuous contact and ricochetal brachiation, for horizontal brachiation with 
regular spacing of handholds~\cite{bertram_point-mass_1999}. 
The solutions are based on circular pendular motion, alternating with parabolic free flight phases
in the case of ricochetal motion.  It is found that natural gibbon motion is even smoother than
the motions predicted by this model, particularly for handhold forces.
Several results successfully developed horizontal continuous brachiation
strategies for a two-link robot~\cite{nakanishi_brachiating_2000} and motions for a three-link model
as demonstrated on a humanoid robot~\cite{kajima_study_2004}.
Zero-energy costs are later demonstrated for a five-link model~\cite{gomes_five-link_2005}.

Methods and results since 2015 have continued to make additional progress.
Non-horizontal ricochetal brachiation is demonstrated for a 2-link primate robot~\cite{wan_non-horizontal_2015}
and uses an analytic solver for the boundary conditions of each swing and a Lyapunov-based tracking controller.
A method is developed for planar brachiation on flexible cables~\cite{farzan_modeling_2018}, for a 3-link model
and corresponding robot. A multiple shooting method is used to plan open-loop feedforward commands.
Brachiation control of a 3-link robot is also designed and demonstrated using iLQR~\cite{yang_design_2019},
for animated sequences of single swings. 
A pendulum model and a planar articulated model are described in the PhD work of Berseth~\cite{Berseth_2019},
based on a finite-state-machine control structure. While demonstrating potential, the results
are based on a "grab anywhere" model and produce controllers with limited capabilities and realism.
The control for a transverse (sideways) ricochetal brachiation has also been investigated recently, for a 4-link arms, body, and tail model~\cite{lin_trbr_2020}.

Our work differs from the bulk of the above work in the following ways.
We demonstrate the ability to learn planar ricochetal brachiation with minimal manual engineering,
and that can traverse handhold sequences with significant variation.
We demonstrate emergent behaviors, such as additional back-and-forth swings as may be needed to proceed in some situations. 
Our approach leverages the benefits of learning with a simplified model,
showing that the simplified-model control policies can provide highly-effective guidance
for learning with the full articulated-body model. The learned control policies anticipate upcoming handholds,
and can be readily integrated with a simple motion planner.

\subsection{Physics-based Character Skills}

Physics-based worlds require simulated characters that are capable of many skills,
including the ability to move through all kinds of environments with speed and agility.
Significant progress has been made for over thirty years, although it is only recently
that we have begun to see approaches that demonstrate scalability, in particular
with regard to being able to imitate a wide variety of motion capture clips with
a physics-based human model, 
e.g.,~\cite{peng2018deepmimic,chentanez2018physics,merel2018neural,park2019learning,drecon2019,won2020scalable,wang2020unicon}.  Recent work has also demonstrated the ability to learn bipedal locomotion
that are capable of navigating difficult step sequences, with learning curricula and 
in the absence of motion capture data~\cite{2020-SCA-allsteps}. 

\subsection{Learning with Simplified Models}

Simplified physical approximations are a standard way to create models that are tractable to simulate and control,
particularly for legged locomotion and brachiation. A common strategy is to first create a motion plan 
using the simplified model, e.g., a spring-loaded inverted pendulum~\cite{poulakakis2009spring}
or centroidal dynamics models, e.g.,~\cite{green_learning_2021, tsounis_deepgait_2020, xie_glide_2021}, and then treat this as a reference motion to be tracked online~\cite{handbook-of-robotics}.
Both the motion planning and tracking problems have commonly been solved using model-based trajectory optimization ({\tt TO}) or model-free reinforcement learning ({\tt RL}) methods. 
In what follows below, we will use a `plan+tracking' notation to loosely categorize combinations of motion planning and tracking of the resulting motion plan. 
There also exists an abundance of literature on pure {\tt TO} and {\tt RL} solutions, which we place outside the scope
of our discussion, although for evaluation purposes, we do consider an RL-only baseline method.

\begin{figure*}
  \centering
  \includegraphics[width=.8\linewidth]{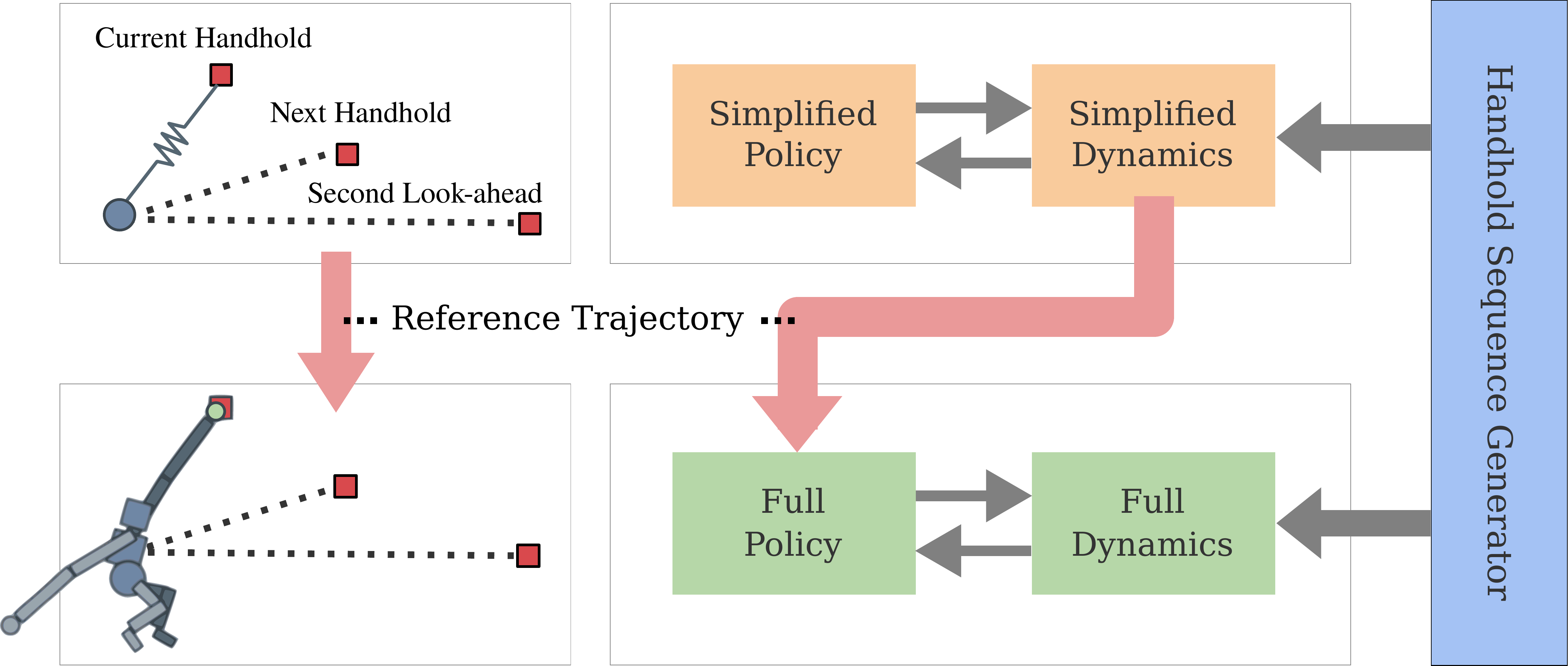}
  \caption{Overview of our simplified model imitation learning system. The simplified model allows for efficient exploration and quickly produces physically-feasible reference trajectories for the full model to imitate.}
  \label{fig:system-diagram}
\end{figure*}

A prominent example of a {\tt TO+TO} approach can be found in the control of the Boston Dynamics Atlas robot~\cite{Kuindersma2020}, where
the planning {\tt TO} uses a long horizon and solves from scratch, and the tracking {\tt TO} 
performs online MPC-style solvers over a short horizon and leverages the planning {\tt TO} results to warm start the optimization.
Another {\tt TO+TO} example first plans legged locomotion with a sliding-based inverted pendulum,
which then initializes a more detailed {\tt TO} optimization with contact locations and a centroidal model~\cite{kwon_fast_2020},
and, lastly, maps this to a full body model using momentum-mapped inverse kinematics.
Much recent work in imitation-based physics-based animation and robotics could be categorized as {\tt fixed+RL}, where captured motion data
effectively serves as a fixed motion plan, e.g.,~\cite{peng2018deepmimic} and related works.

Quadruped control policies have been developed using both {\tt TO+RL} and {\tt RL+TO} approaches, 
e.g., \cite{RL-adapt} and \cite{RLOC,xie_glide_2021}, respectively.
A recent {\tt RL+RL} approach uses a simplified model (and related policy) that is allowed to sample in state space rather
than action space~\cite{two-stage-DRL}.  This allows it to behave more like a motion planner and assumes that the result can
still provide a useful motion plan to for the full model to imitate. 
The method demonstrates the ability to avoid a stand-in-place local minima for a simulated quadruped, which  may arise due to sparse rewards.
Our work develops an {\tt RL+RL} approach for the challenging control problem of brachiation.
In our case, both the simplified and full MDPs work directly in a physically-grounded action space.
We demonstrate the benefit of this two-level approach via multiple baselines and ablations.

\section{Environments}

\subsection{Simplified Model}
\label{sec:env-simplified-model}

Our simplified model treats the gibbon as a point mass equipped with a virtual extensible arm.
The virtual arm is capable of grabbing any handhold within the annular area defined by a radius $r \in [r_{\min},r_{\max}]$,
with $r_{\min}$=10 cm, and $r_{\max}$=75 cm.
We note that while the $r_{\max}$ is greater than the arm length of the full model (60~cm), the simplified model is intended to be a proxy for the center of mass of the gibbon, and not the shoulder.

The simplified model dynamics consist of two alternating phases.
The swing phase starts when the character performs a grab when one of its hands is at (or nearby) to the target handhold.
During this phase, the character can apply a force along the direction of the grabbing arm, either to pull towards or push away from the current handhold.
The swing phase dynamics is equivalent to a spring-and-damper point-mass pendulum system.
The character can influence the angular velocity by shortening or lengthening the swing arm of the pendulum.
The flight phase is defined as the period when neither hand is grabbing.
During this phase, the character's trajectory is governed by passive physics, following a parabolic trajectory.
The control authority that remains during a flight phase comes from the decision
of when to grab onto the next handhold, if it is within reaching distance.
The control parameters are described in detail in \autoref{sec:action-space}.

The minimum and maximum arm length are enforced by applying a force impulse when the length constraint is violated during a swing phase.
We implement the physics simulation of the simplified environment in PyTorch~\cite{paszke2019pytorch}, which can be trivially parallelized on a GPU.

\subsection{Full Model}

The full model is a planar articulated model of a gibbon-like character 
that approximates the morphology of real gibbons, as reported in \cite{michilsens_functional_2009}, and is simulated using PyBullet~\cite{pybullet}. 

Our gibbon model consists of 13 hinge joints including shoulders, elbows, wrists, hips, knees, ankles, and a single waist joint at the pelvis.
All joints have associated torques, except for the wrist joints, which we consider to be passive.
To capture the 3D motion of the ball-and-socket joints in two dimensions, we model the shoulder joints as hinge joints without any corresponding joint limits.
This allows the simulated character to produce the \textit{under-swing} motions
of real gibbons.
Our gibbon has a total mass of 9.8~kg, an arm-reach of 60~cm, and a standing height of 63 centimeters.
We artificially increase the mass of the hands and the feet to improve simulation stability.
The physical properties of our gibbon model are summarized in \autoref{tab:gibbon-summary}. 

\begin{table}
\centering
\small
\caption{\label{tab:gibbon-summary}Physical Properties of Full Gibbon Model. The waist joint link mass and length includes the torso and the head.  Others include only the direct child link.}
\begin{tabular}{@{}lccc@{}}
\toprule
Joint & Max.~Torque (Nm) & Link Mass (kg) & Link Length (cm)\\
\midrule
Waist & 21 & 3.69 & 31\\
Shoulder & 35 & 0.74 & 26\\
Elbow & 28 & 0.74 & 26\\
Wrist & 0 & 0.40 & 8\\
Hip & 14 & 0.37 & 12\\
Knee & 14 & 0.31 & 12\\
Ankle & 7 & 0.47 & 8\\
\bottomrule
\end{tabular}
\end{table}

The grab behavior in the full model is simulated using point-to-point constraints.  
This allows us to abstract away the complexity of modelling hand-object dynamics, yet still provides a reasonable approximation to the underlying physics.
In our implementation, a point-to-point constraint is created when the character performs a grab and the hand is within five centimeters of the target handhold.
The constraint \textit{pins} the grabbing hand in its current position, as opposed to the target handhold location, to avoid introducing undesirable fictitious forces.
At release, the constraint is removed.

\subsection{Handhold Sequences Generation}
\label{sec:sequence-generation}

Both the simplified and full characters operate in the same environment, which contains handhold sequences. In a brachiation task, the goal is to grab successive handholds precisely and move forward.
Successive handholds are generated from uniform distributions defined by the distance, $d \sim U(1,2)$~meters, and pitch, $\phi \sim [-15^{\circ},15^{\circ}]$, relative to the previous handhold.

\subsection{Action Spaces}
\label{sec:action-space}

The simplified and full models both use a control frequency of 60 Hz and a simulation frequency is 480 Hz.

The action space for the simplified model consists of two values, a target length offset and a flag indicating to grab or to release.
At the beginning of each control step, the target length offset is added to the current arm length to form the desired target length.
In addition, the grab flag is used to determine the current character state.
At each simulation step, the applied force is computed using PD control based on the desired target length.
We clamp the maximum applied force to 240 N, close to the maximum observed forces in real gibbon reported in \cite{michilsens_functional_2009}.

In the full model, the action space is analogous to that of the simplified model, only extended to every joint.
An action consists of 15 parameters representing the joint angle offsets for each of the 13 joints and a flag indicating grab or release for each hand.
The applied torques in each joint are computed using PD control.
We set the proportional gain, $k_p$, for each joint to be equal to its range of motion in radians and the derivative gain to be $k_p / 10$.

\subsection{State Spaces}
\label{sec:state-space}

Both environments have a similar state space, composed of information of the character and information on the future handholds sequence to grab.

In the simplified environment, the character state is three dimensional consisting of the root velocity in world coordinates and the elapsed time since the start of the current swing phase.
The elapsed time information is necessary to make the environment fully observable since the environment enforces a minimum and a maximum grab duration (\hyperref[sec:termination-conditions]{\S\ref*{sec:termination-conditions}}).
We normalize the elapsed time by dividing by the maximum allowable grab duration.
A zero indicates that the character is not grabbing and a one indicates it has grabbed for the maximum amount of time.

The character state for the full model is a 45D vector consisting of root velocity, root pitch, joint angles, joint velocities, grab arm height, and grab states.
The torso is used as the root link.
The root velocity ($\mathbb{R}^2$) is the velocity in the sagittal plane.
Joint angles ($\mathbb{R}^{26}$) are represented as $\cos(\theta)$ and $\sin(\theta)$ of each joint, where $\theta$ is the current angle.
Joint velocities ($\mathbb{R}^{13}$) are the angular velocities measured in radians per second.
Grab arm height ($\mathbb{R}$) is the distance between the free hand and the root in the upward direction, which can be used to infer whether the next target is reachable.
Lastly, the grab states ($\mathbb{R}^2$) are Boolean flags indicating if each hand is currently grabbing.

In both environments, the control policy receives task information pertinent to the handhold locations in addition to the character state.
In particular, the task information includes the location of the current handhold and the $N$ upcoming handholds in the character's coordinate frame. 
We experiment with different values of $N$ in \autoref{sec:results-simplified-policy}.

\section{Learning Control Policies}

We use deep reinforcement learning (DRL) to learn brachiation skills in both the simplified and the detailed environment. 
In RL, at each time step $t$, the control policy reacts to an environment state $s_t$ by performing an action $a_t$.
Based on the action performed, the policy receives a reward signal $r_t = r(s_t, a_t)$ as feedback. 
In DRL, the policy computes $a_t$ using a neural network $\pi_\theta(a | s)$, where $\pi_\theta(a | \cdot)$ is the probability density of $a$ under the current policy. 
The goal of DRL is to find the network parameters $\theta$ which maximize the following:
\begin{align}
J_{RL}(\theta) = E\left[\sum_{t=0}^{\infty}\gamma^t{r({s}_t, {a}_t)} \right].
\end{align}
Here, $\gamma \in [0, 1)$ is the discount factor so that the sum converges.
We solve this optimization problem using the proximal policy optimization (PPO) algorithm~\cite{PPO}, a policy gradient actor-critic algorithm.
We choose PPO for its effective utilization of hardware resources in parallelized settings which results in reduced wall-clock learning time.
Our learning pipeline is based on a publicly available implementation of PPO \cite{pytorchrl}. A review of the PPO algorithm is provided in 
~\autoref{appendix:ppo-summary}.

\subsection{System Overview}

We start by providing an overview of our system, shown in \autoref{fig:system-diagram}.
Our system contains three distinct components: the simplified model, the full model, and a handhold sequence generator.
During training, the simplified model is trained first on randomly sampled handhold sequences generated by the handhold generator.
After the simplified model is trained, we obtain reference trajectories from the simplified environment on a fixed set of handholds sequences.
The full model is subsequently trained by imitating the reference trajectories from the simplified model while optimizing task and style objectives.
At evaluation time, the simplified model can be used as a planner to provide guidance for the full model either upfront for an entire handhold sequence or on a per-grab basis. 
Both models working in tandem allows our system to traverse difficult handholds sequences that would otherwise be challenging for standard reinforcement learning policies.

\subsection{Learning Simplified Policies}
\label{sec:learning-simplified-policies}

The simplified policy is trained to optimize a sparse reward that is given when the character grabs the next handhold.
The fact that the simplified policy can be trained with only the sparse task reward is desirable as it allows for interesting behaviors to emerge.
In DRL, reward shaping is often required for the learning to succeed or to significantly speed up the learning.
At the same time, reward shaping can also introduce biases that cause the control policy to optimize for task-irrelevant objectives and deviate from the main task.
Directly optimizing a policy using the sparse task reward allows us to find solution modes that would otherwise be exceedingly difficult to find.

\paragraph{Simplified Policy Networks}
The simplified policy contains two neural networks, the controller and the value function.
The controller and the value function have similar architecture differing only in the output layer.
Each neural network is a three-layered feed-forward network with 256 hidden units followed by ReLU activations.
Since the actions are to be scaled by the proportional gain constants in the PD controller, we apply Tanh activation to the controller network's output layer to ensure the values are normalized between -1 and +1.
The value function outputs a scalar value approximating the expected return which is not normalized.
The input to the networks are as described earlier~(\hyperref[sec:state-space]{\S\ref*{sec:state-space}}).

\begin{figure*}
  \begin{subfigure}[t]{0.30\textwidth}
    \centering
    \includegraphics[width=\textwidth]{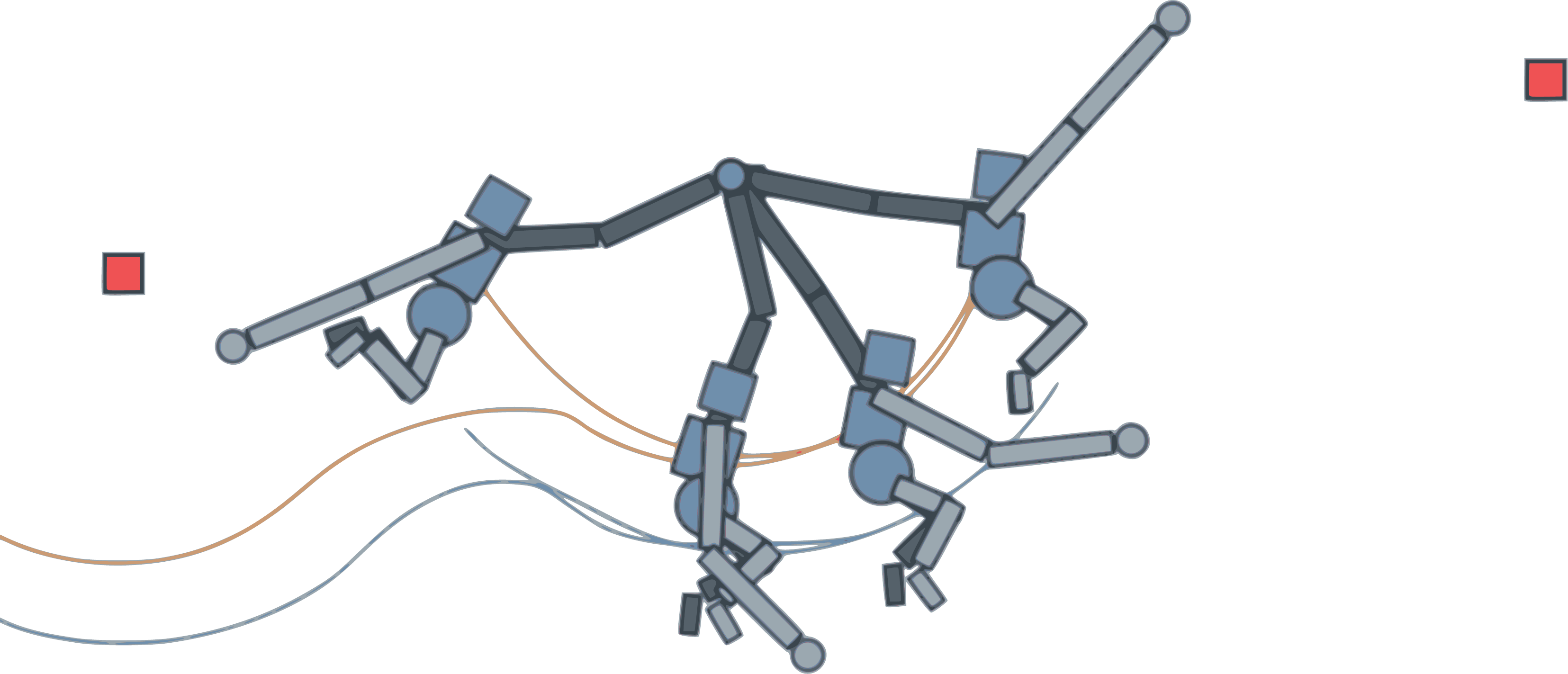}
    \caption{Backward swing}
  \end{subfigure}
  \begin{subfigure}[t]{0.30\textwidth}
    \centering
    \includegraphics[width=\textwidth]{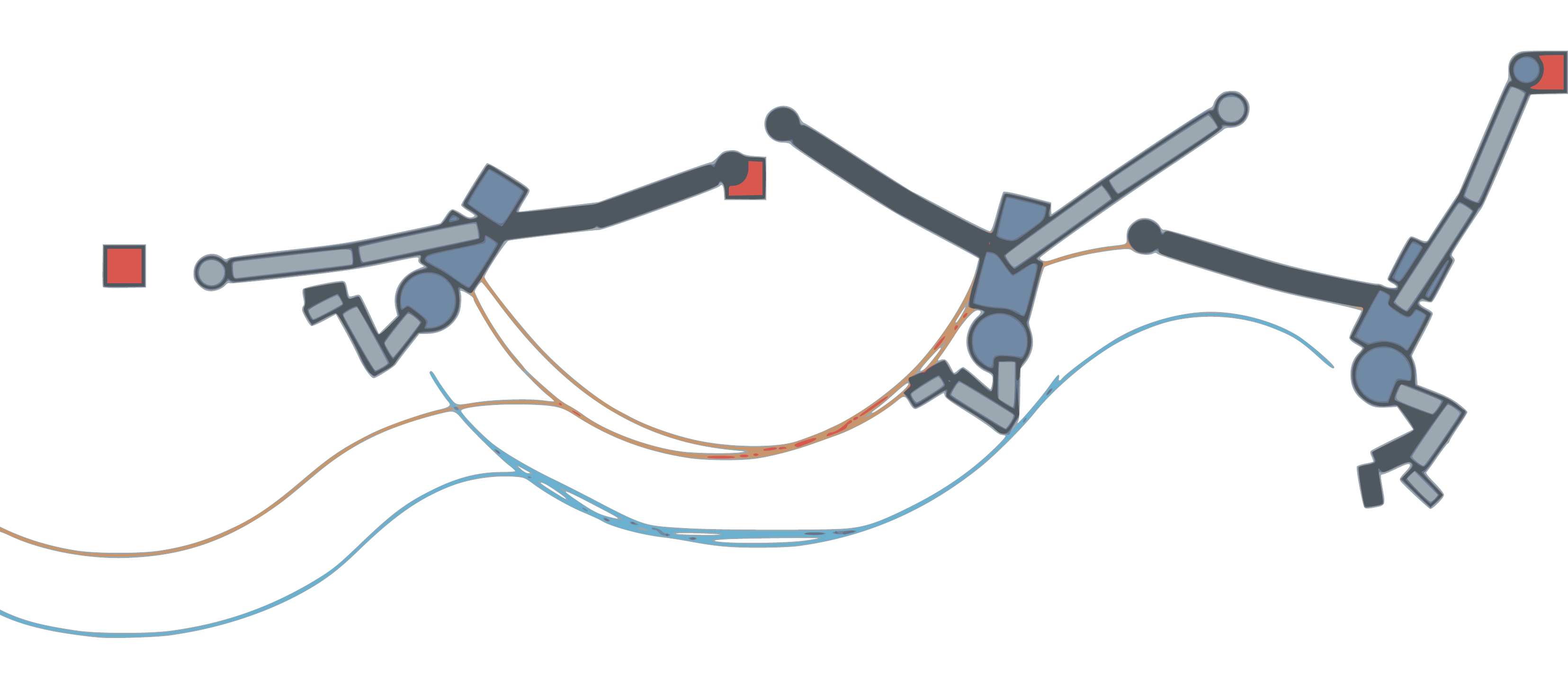}
    \caption{Forward swing}
  \end{subfigure}
  \begin{subfigure}[t]{0.30\textwidth}
    \centering
    \includegraphics[width=\textwidth]{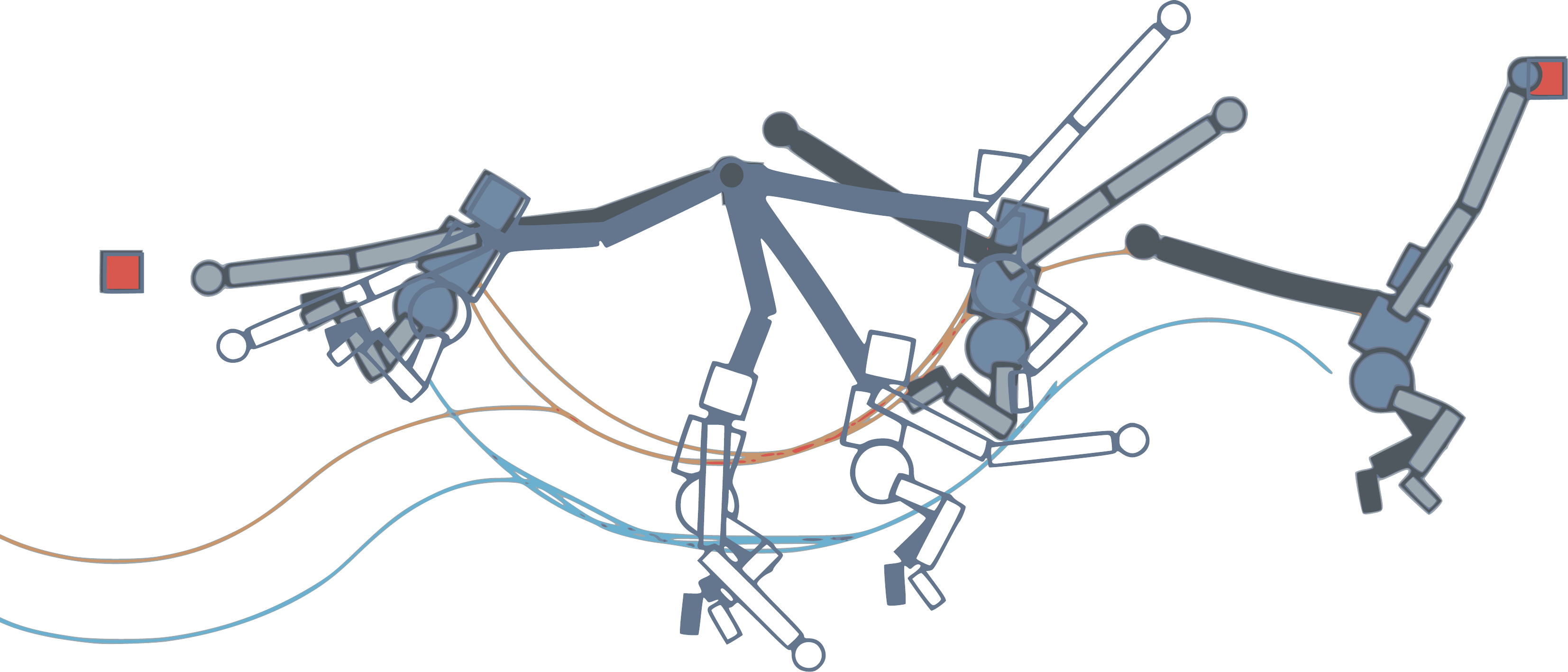}
    \caption{Complete {\em pumping} trajectory}
  \end{subfigure}
  \caption{Brachiation with emergent {\em pumping behavior}. 
  The blue line gives the trajectory of the simplified model,
  while the brown line gives the trajectory of the root of the full character.
  The disparity between the two is caused by the fact that the simplified model has a longer arm length with respect to the full model, to better represent its center of mass. For further visual trajectories please refer to
  \autoref{appendix:supplementary-results} and to the supplementary video.}
  \label{fig:motions}
\end{figure*}

\subsection{Learning Full Model Policies Using Simplified Model Imitation}
\label{sec:simplified-model-imitation}

The full model policy is trained using imitation learning by tracking the reference trajectory generated by the simplified model.
The imitation learning task can be considered as an inverse dynamics problem where the full model policy is required to produce the joint torques at each time step given the current character state and the future reference trajectory.
We can consider the reward function to have a task reward component, an auxiliary reward component, and a style reward component.

As in the simplified environment, the task reward, $r_\text{task}$, is the sparse reward for successfully grabbing the next target handhold.
The auxiliary reward ($r_\text{aux}$) consists of reward objectives that facilitate learning, including a tracking term and a reaching term.
The tracking term rewards the character to closely follow the reference trajectory and a reaching term encourage the grabbing hand to be close to the next target handhold.
The style reward ($r_\text{style}$) terms are added to incentivize more natural motions, including keeping the torso upright, slower arm rotation, reducing lower body movements, and minimizing energy. 
Finally, the overall reward can be computed as:
\begin{align}
    r_\text{full} &= \exp(r_\text{aux} + r_\text{style}) + r_\text{task}\\
    r_\text{aux} &= w_{t} r_\text{tracking} + w_{r} r_\text{reaching}\\
    r_\text{style} &= w_{u} r_\text{upright} + w_{a} r_\text{arm} + w_{l} r_\text{legs} + w_{e} r_\text{energy}
\end{align}
Note that the style reward does not affect the learning outcome; the character can learn brachiation motions without the style reward.
The reward terms and their weighting coefficients are fully described in
\autoref{appendix:full-model-rewards}.

\paragraph{Full Policy Networks}
The controller and the value function of the full model consist of two five-layer neural networks,
with a layer width of 256. The first three hidden layers of the policy use the softsign
activiation, while the final two hidden layers use ReLU activation. A Tanh is applied to the final output to normalize the actions.  The critic uses ReLU for all hidden layers.
Prior work is contradictory in nature with respect to the impact of network size, finding larger network sizes to perform both better \cite{wang2020unicon} and worse \cite{drecon2019} for full-body motion imitation tasks.
In our experiments, we find the smaller networks used for the simplified model to be incapable of learning the brachiation task in the full environment.

\subsection{Initial State and Early Termination}
\label{sec:termination-conditions}
For both environments, we initialize the character to be hanging from the first handhold at a 90-degree angle, where zero degree corresponds to the pendulum resting position.
The initial velocity of the character is set to zero as the base state provides the necessary forward momentum to reach the next handhold.

The environment resets when the termination condition is satisfied, which we define as entering an unrecoverable state.
An unrecoverable state is reached when the character is not grabbing onto any handholds and falls below the graspable range of the next handhold with a downward velocity.

In addition to the unrecoverable criteria, both environments implement a minimum and maximum grab duration.
The minimum and maximum duration are set to 0.25 and 4 seconds respectively.
The idea of the minimum grab duration is to simulate the reaction time and time required to physically form a fist during a grasp.
The maximum grab duration effectively serves as another form of early termination.
It forces the character to enter an unrecoverable state unless the policy has already learned to generate forward momentum.
We check the importance of early termination in \autoref{sec:results}.
\section{Results}
\label{sec:results}

All experiments are performed on a single 12-core machine with one NVIDIA RTX 2070 GPU.
For the simplified model, training takes approximately five minutes as the simulation is parallelized on the GPU.
Training the full model policy takes just under one hour.
Each experiment is limited to collect a total of 25 million samples and the learning rate is decreased to $3 \times 10^{-5}$ over time using exponential annealing.
We use the Adam~\cite{kingma2014adam} optimizer with a mini-batch size of 2000 and a learning rate of $3 \times 10^{-4}$ to update policy parameters in all experiments.
Other implementation and hyperparameter details are summarized
in \autoref{appendix:ppo-hyperparameters}.


\subsection{Simplified Model Results}
\label{sec:results-simplified-policy}

The goal of the simplified model is to facilitate more efficient training of a full model policy by generating physically-feasible reference trajectories.
A baseline solution for the simplified model can be successfully trained to find reasonable solutions for traversing difficult terrains, i.e. with handholds sampled from the full distribution (\shortsectionref{sec:sequence-generation}), using only the task reward described in \autoref{sec:learning-simplified-policies}.
The learned controllers have some speed variations; we use a low speed controller to train the full model policy as the generated trajectories are closer to the motions demonstrated by real gibbons.


The learned control policies can traverse challenging sequences of handholds which require significant flight phases and exhibit emergent \textit{pumping} behavior where the characters would perform extra back-and-forth swings as needed.
We observe three scenarios where pumping behavior can be observed: waiting for the minimum swing time to elapse, adjusting for a better release angle, and for gaining momentum.
A visualization of the pumping behavior is shown in~\autoref{fig:motions}.
Please refer to the supplementary video for visual demonstrations of generated trajectories.

\paragraph{Number of Look-ahead Handholds}


In this experiment, we empirically verify the choice for the number of look-ahead handholds by comparing the task performance on $N=\{1,2,3,5,10\}$.
Results are summarized in \autoref{tab:ablation-num-lookahead}.
The best performance is achieved when $N=1$ as measured by both the number of completed handholds and the average episode reward.
This is surprising since previous work in human locomotion has shown that a two-step anticipation horizon achieves the lowest stepping error~\cite{coros2008_constrained_walking}.
The result shows that the policy performance generally decreases with increasing number of look-ahead handholds.
We hypothesize that, due to the relative small network size used in the simplified policy, increasing the number of look-ahead handholds creates more distraction for the policy.
We leave validation of this hypothesis for future work.

\paragraph{Importance of Early Termination}

In this ablation study, we verify the importance of applying early termination in the simplified environment.
There are two forms of early termination: the unrecoverable criteria and the maximum grab duration.
The complete simplified model implements all early termination strategies.
\autoref{tab:ablation-termination} shows the average number of handholds reached aggregated across five different runs.
The results show that enforcing strict early termination can facilitate faster learning.
When maximum grab duration is disabled, only one run out of ten total successfully learned the brachiation task.
In addition, we find that enforcing the minimum grab duration of 0.25 seconds helps the character to learn to swing, which in turn helps to discover the future handholds.

\begin{table}
\renewcommand{\arraystretch}{1.2}
\centering
\small
\caption{\label{tab:ablation-num-lookahead}Experiment results on the effect of the number of look-ahead handholds for the simplified model.  We report the average and the standard deviation over five independent runs.}
\begin{tabular}{@{}ccccc@{}}
\toprule
\multicolumn{5}{c}{\textit{Number of Look-ahead Handholds}}\\
1 & 2 & 3 & 5 & 10\\
\midrule
\multicolumn{5}{l}{\textit{Handholds Completed}}\\
$\mathbf{17.8 \boldsymbol\pm 2.1}$ & $14.6 \pm 5.0$ & $4.9 \pm 5.7$ & $7.6 \pm 5.8$ & $0.3 \pm 0.6$\\
\midrule
\multicolumn{5}{l}{\textit{Episode Reward ($\times 10^2)$}}\\
$\mathbf{7.3 \boldsymbol\pm 0.8}$ & $5.9 \pm 1.8$ & $2.0 \pm 2.4$ & $3.1 \pm 2.4$ & $0.1 \pm 0.3$\\
\bottomrule
\end{tabular}
\end{table}

\begin{table}
\renewcommand{\arraystretch}{1.2}
\setlength{\tabcolsep}{0.8\tabcolsep}
\centering
\small
\caption{\label{tab:ablation-termination}Ablation results on early termination strategies for the simplified model. Each column names the termination strategy that is removed.  Each entry represents the average number of completed handholds and the standard deviation over five runs.}
\begin{tabular}{@{}ccccc@{}}
\toprule
\multicolumn{5}{c}{\textit{Early Termination Ablations}}\\
None & Unrecoverable & Min Duration & Max Duration & Min \& Max\\
\midrule
$\mathbf{14.6 \boldsymbol\pm 5.1}$ & $10.2 \pm 7.3$ & $7.4 \pm 9.3$ & $2.9 \pm 2.0$ & $3.1 \pm 4.1$\\
\bottomrule
\end{tabular}
\end{table}




\subsection{Full Model Results}
\label{sec:full-model-results}

The full model policy is learned through simplified model imitation, as described 
earlier~(\hyperref[sec:simplified-model-imitation]{\S\ref*{sec:simplified-model-imitation}}). 
In addition to using the tracking reward, which uses the simplified model only during training, we experiment with three different methods of further leveraging the simplified model reference trajectories at inference time.
\begin{itemize}[leftmargin=\parindent]
    \setlength\itemsep{3pt}
    \item[] \textbf{A. Tracking reward only}. The tracking reward is combined with the task reward to train the full model policy.  The reference trajectory is only used to compute the tracking reward.
    \item[] \textbf{B. Tracking + Other Rewards}. Reward includes all reward terms: tracking, task, auxiliary, and style rewards.
    \item[] \textbf{C. Rewards + Release Timing}. Reward includes everything described in Experiment B.  In addition, the release timing from the reference trajectory is used to control the grab action of the full model policy.
    \item[] \textbf{D. Rewards + States}. Reward includes everything described in Experiment B. In addition, a slice of the future reference trajectory is included as a part of the full model policy input.
    \item[] \textbf{E. Rewards + States + Grab Info}. Reward includes everything described in Experiment D. In addition, instead of conditioning the full model state with just the future reference trajectory, we also give access to the grab flag for those future points.
\end{itemize}

\begin{figure}
  \centering
  \includegraphics[width=.95\linewidth]{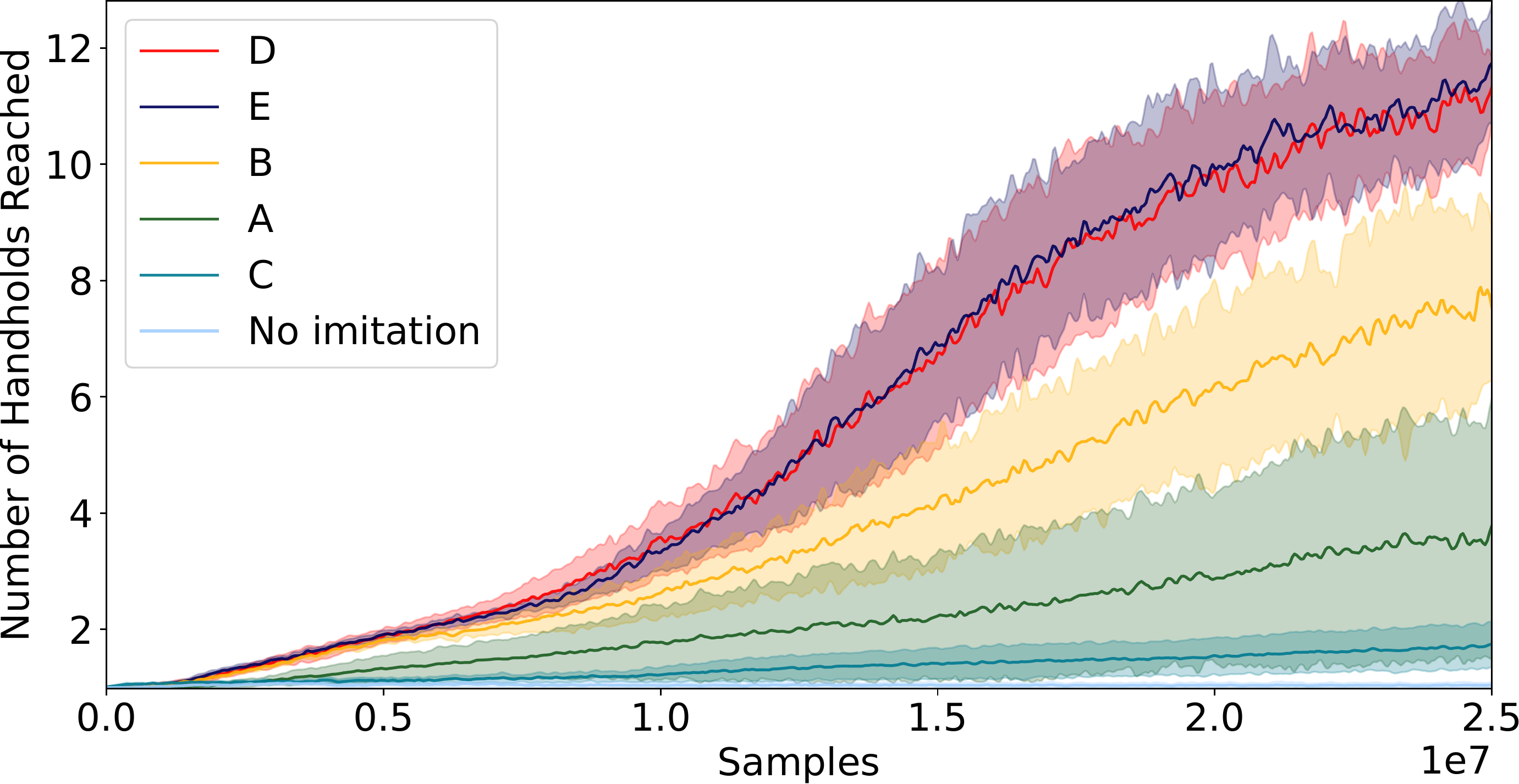}
  \caption{Learning Curves for the full model, for various configurations.
   No imitation reward results in failure. 
   Using reference trajectory information only at training time (A and B) improves learning.
   Using the release timing from the simplified model degrades performance (C).
   The best policies are obtained when simplified and full models are used in tandem at evaluation time (D and E).
  Please refer to \autoref{sec:full-model-results} for a detailed description.}
  \label{fig:ablation-full-model}
\end{figure}

\autoref{fig:ablation-full-model} shows the learning curves for different full model configurations.
We include a baseline model which is trained using all reward terms except for the tracking reward. 
For the baseline model, we experimented both with and without a learning curriculum on the terrain difficulty, e.g., similar to~\cite{2020-SCA-allsteps}, but neither version was able to advance to the second handhold.
The best policies are obtained when the simplified model is also used at evaluation time.
This includes adding information from the reference trajectory (Experiment~D) and also further adding the grab information (Experiment~E).
However, the grab information provides no additional benefit to the policy when reference trajectory information is present.

\subsection{Full Model with Planning}

\begin{table}
    \centering
    \caption{Quantitative evaluation of the different planning methods. In each generated terrain, three unreachable gaps (e.g., \autoref{fig:planning-visuals}) are placed at random locations. Each planning method is evaluated on 40 different terrains. The combined approach is able to pass all gaps most consistently.}
    \begin{tabular}{lcccc}
    \toprule
    & \multicolumn{4}{c}{Gaps Passed}\\
    Planning Method & 0 & 1 & 1+2 & 1+2+3 \\
    \midrule
    Only Full Value Function & 40 & 22 & 9 & 7\\
    Only Simplified Rewards & 40 & 36 & 26 & 15\\
    Reward \& Value Function & 40 & 35 & 26 & \textbf{21}\\
    \bottomrule
    \end{tabular}
    \label{tab:planning-results}
\end{table}

\begin{figure*}
  \begin{subfigure}[t]{\columnwidth}
    \centering
    \includegraphics[width=.8\columnwidth]{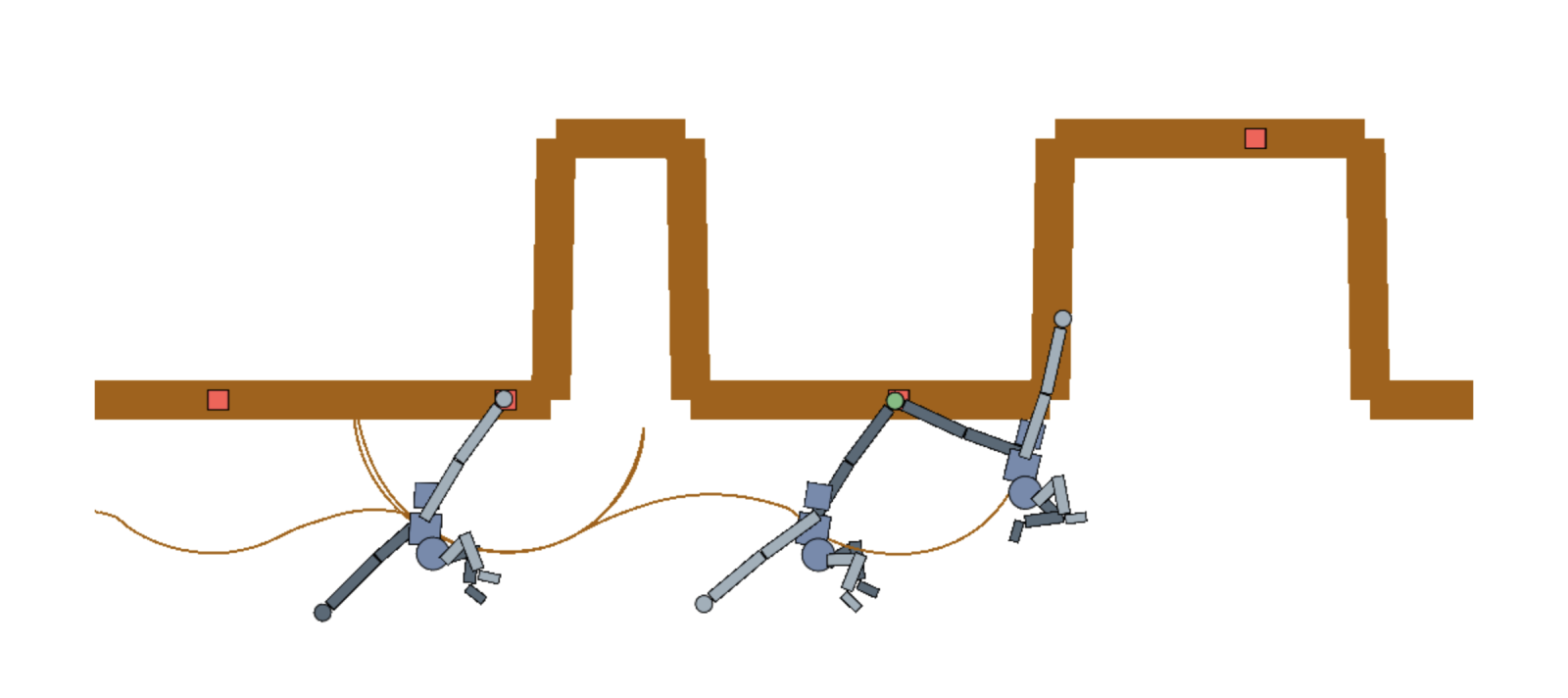}
    \captionsetup{margin=0.025\columnwidth}
    \caption{Planning using only the full model value function. The last planned target grasp location (red dot) is unreachable. This approach has a limited planning horizon, so unrealistic plans are occasionally generated.}
  \end{subfigure}
  \begin{subfigure}[t]{\columnwidth}
    \centering
    \includegraphics[width=.8\columnwidth]{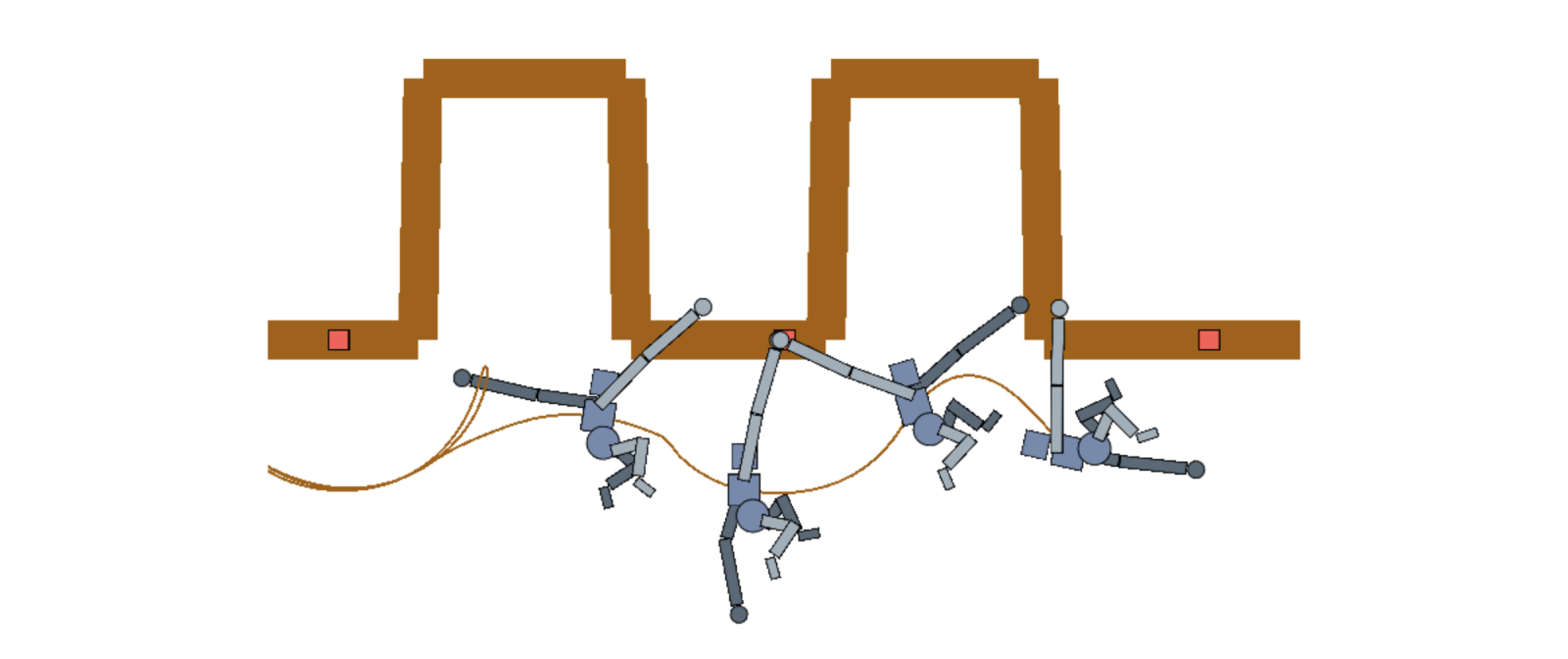}
    \captionsetup{margin=0.025\columnwidth}
    \caption{Planning using only the simplified model. Here, the last handhold target is unreachable because this planning approach fails to consider the capability of the full model policy.}
  \end{subfigure}
  \begin{subfigure}[t]{1.6\columnwidth}
    \centering
    \includegraphics[width=0.5\columnwidth]{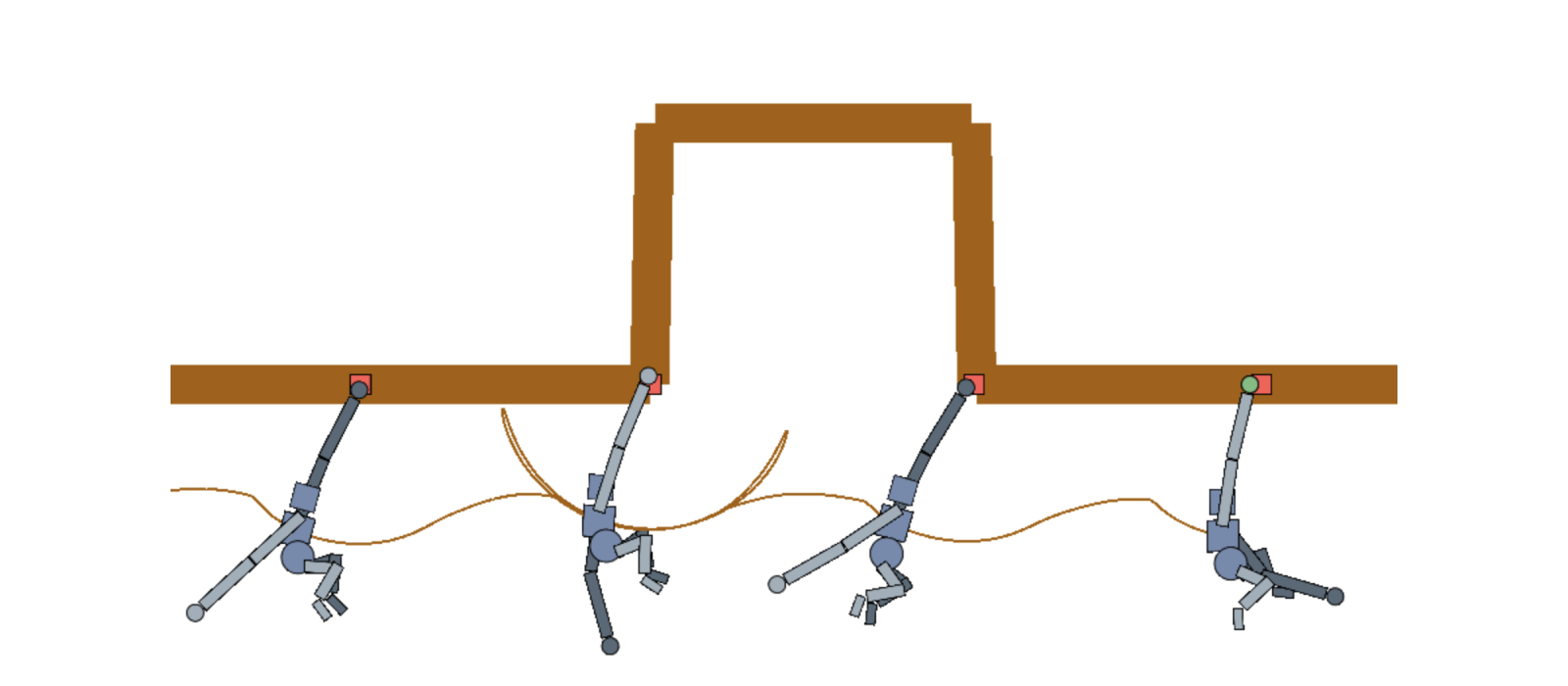}
    \caption{Planning with combined full model value function and simplified model. The target handhold locations are planned very close to the discontinuities, which effectively reduces the risk of producing unreachable targets.}
  \end{subfigure}
  \caption{Comparison of different planning approaches.}
  \label{fig:planning-visuals}
\end{figure*}

At inference time, we are able to extend the capability of the full model policy by expanding the anticipation horizon using the simplified model.
The full model policy's value function cannot anticipate more than the number of look-ahead handholds that was used for training.
However, the simplified model is fast enough to replan on a handhold-by-handhold basis, in the style of model predictive control (MPC).
It can be used to simulated thousands of trajectories in parallel at run-time.
In this experiment, we set the planner to explore 10k trajectories for 4 seconds into the future; this can be completed in under one second of machine time.
There is still room for improvement for real time animation purposes.


The planner uses three fully-trained components: the full model policy $\pi_\text{full}$, the full model value function $V_\text{full}$ and the simplified model policy $\pi_\text{simple}$.
As input, the method takes a terrain which is a contiguous ceiling where handholds can be placed.
Given the current state of the full model character $s_t$, we randomly sample $K$ handhold plans of length $H$ handholds on the terrain, $\{\mathcal{P}_k\}$.
Concretely, each handhold in a plan $\mathcal{P}_k$ is sampled at a horizontal distance $\delta x \sim U(1.1,1.8)$ beyond the previous handhold,
and takes its corresponding height from the terrain.
We then retain the best plan, i.e.,  $k = \arg\max J(\mathcal{P}_k)$, according to the objective defined by
\begin{align*}
  J(\mathcal{P}_k) = V_\text{full}(s_t, k[0:N]) + \sum_j^H R_j,
\end{align*}
where $R_j$ is the reward for the simplified model obtained by $\pi_\text{simple}$, and $N$ is the policy look-ahead value, with $H>N$.
The selected plan is then used until the next handhold is reached, at which point a full replanning takes place. This MPC-like process repeats \textit{ad infinitum}.
We find empirically that the above objective generates better planning trajectories as compared to only using the reward of $\pi_\text{simple}$ or only $V_\text{full}$.
This is because the reward of the simplified model is not enough to capture the capabilities of the full character, while the value function has a limited planning horizon.
Quantitative results are presented in \autoref{tab:planning-results}.
Examples of terrains traversed by the policy with the planners are presented in \autoref{fig:planning-visuals} and in the supplementary video.
\section{Conclusions}

Brachiation is challenging to learn, as it requires the careful management
of momentum, as well as precision in grasping. 
We demonstrate a two-level reinforcement learning method
for learning highly-capable and dynamic brachiation motions.
The motions generated by the policy learned for the simplified model 
play a critical role in the efficient learning of the full policy.
Our work still has many limitations. There are likely to be other
combinations of rewards, reward curricula, and handhold curricula, that,
when taken together, may produce equivalent capabilities.
We have not yet fully characterized the limits of the model's capabilities,
i.e., the simulated gibbon's ability to climb, leap, descend, and more.
We wish to better understand, in a quantitative fashion, how the 
resulting motions compare to those of real gibbons. 
Lastly, to emulate the true capabilities of gibbons, we need methods
that can plan in complex 3D worlds, perform 3D brachiation, make
efficient use of the legs when necessary, and more.
We believe the simplicity of our approach provides a solid foundation for future research in these directions.

\begin{acks}
This work was supported by the NSERC grant RGPIN-2020-05929.
\end{acks}

\bibliographystyle{ACM-Reference-Format}
\clearpage
\bibliography{bibliography}

 \clearpage
\appendix
\section{Full Model Rewards}
\label{appendix:full-model-rewards}

The tracking reward $r_\text{tracking}$ is formulated as a penalty and computed as the squared distance between the character's root and the position of the reference trajectory.
\begin{align}
    r_\text{tracking} &= \| p_\text{body} - p_\text{reference} \|^2_2 \nonumber\\
    &\;w_t = -4 \nonumber
\end{align}

The reaching reward $r_\text{reaching}$ is computed similarly as the tracking reward, except the distance is between the next grabbing hand and the target handhold.
This reward is only applied during the flight phase.
During swing phase, the character can still use the free arm to generate momentum without penalty.
\begin{align}
    r_\text{reaching} &= \| p_\text{hand} - p_\text{target} \|^2_2, \; \text{if in flight phase} \nonumber\\
    &\;w_r = -0.1 \nonumber
\end{align}

The upright posture term $r_\text{upright}$ penalizes the character if the root pitch is not within 40 degrees of the vertical axis.
\begin{align}
    r_\text{upright} &= |\text{pitch}| - 40^\circ, \; \text{if $|\text{pitch}| > 40^\circ$} \nonumber\\
    &\;w_u = -1 \nonumber
\end{align}

The arm rotation reward $r_\text{arm}$ penalizes the character for excessive angular velocity in the next grabbing arm.
\begin{align}
    r_\text{arm} &= |\omega_\text{arm}| \text{ and } w_a = -0.1 \nonumber
\end{align}

The lower body reward $r_\text{legs}$ encourages the knees to be close to 110 degrees from base position of straight legs.
This reward is applied to improve the visibility of the legs during brachiation.
\begin{align}
    r_\text{legs} &= \|\theta_\text{knees} - 110^\circ\|_1 \nonumber\\
    &\;w_l = -0.1 \nonumber
\end{align}

The energy reward $r_\text{energy}$ penalizes the policy for using excessive amounts of energy.
The overall energy is approximated based on the applied torques and the combined angular velocity of all joints, computed as:
\begin{align}
    r_\text{energy} &= \|\tau_\text{joints}\|^2_2 + \|\omega_\text{joints}\|_1 \nonumber\\
    &\;w_l = -0.01 \nonumber
\end{align}

\section{Proximal Policy Optimization}
\label{appendix:ppo-summary}
Let an experience tuple be $e_t = (o_t, a_t, o_{t+1}, r_t)$ and a trajectory be $\tau = \{e_0, e_1, \dots, e_T\}$. 
We episodically collect trajectories for a fixed number of environment transitions and we use this data to train the controller and the value function networks.
The value function network approximates the expected future returns of each state, and is defined for a policy $\pi$ as
\begin{align*}
V^\pi(o) = E_{o_0=o, a_t\sim \pi(\cdot | o_t)}\left[\sum_{t=0}^{\infty}\gamma^{\;t} r(o_t, a_t) \right].
\end{align*}
This function can be optimized using supervised learning due to its recursive nature:
\begin{align*}
    V^{\pi_{\theta}}(o_t) = \gamma\;V^{\pi_{\theta}}(o_{t+1}) + r_t,
\end{align*}
where
\begin{align*}
    V^{\pi_{\theta}}(o_T) = r_T + \gamma V^{\pi_{\theta_{old}}}(o_{T+1}).
\end{align*}
In PPO, the value function is used for computing the advantage
\begin{align*}
    A_t = V^{\pi_\theta} - V^{\pi_{\theta_{old}}}
\end{align*}
which is then used for training the policy by maximizing:
\begin{align*}
    L_{\pi}(\theta) = \frac{1}{T}\sum_{t=1}^T \min(\rho_t\hat{A}_t, \; \text{clip}(\rho_t,1-\epsilon,1+\epsilon)\hat{A}_t),
\end{align*}
where $\rho_t = \pi_{\theta}(a_t | o_t) \mathbin{/} \pi_{\theta_{old}}(a_t | o_t)$ is an importance sampling term used for calculating the expectation under the old policy $\pi_{\theta_{old}}$. 

\section{PPO Hyperparameters}
\label{appendix:ppo-hyperparameters}

\begin{table}[H]
\centering
\small
\caption{Hyperparameters used for training PPO.}
\begin{tabular}{lc}
\toprule
\bf{Hyperparameter} & \bf{Simplified/Full Model} \\
\midrule
Learning Rate & $3 \times 10^{-4}$ \\
Final Learning Rate & $3 \times 10^{-5}$ \\
Optimizer & Adam \\
Batch Size & $2000$ \\
Training Steps & $2.5 \times 10^7$ \\
Num processes & $10000$/$125$ \\
Episode Steps & $80000$/$40000$ \\
Num PPO Epochs & $10$ \\
Discount Factor $\gamma$ & $0.99$ \\
Gradient Clipping & False \\
Entropy Coefficient & $0$ \\
Value Loss Coefficient & $0$ \\
Clip parameter & $0.2$ \\
Max Grad Norm & $2.0$ \\ 
\bottomrule
\end{tabular}
\end{table}

\begin{figure*}
\section{Supplementary Results}
\label{appendix:supplementary-results}
\vspace{4pt}
  \centering
  \includegraphics[width=.75\textwidth]{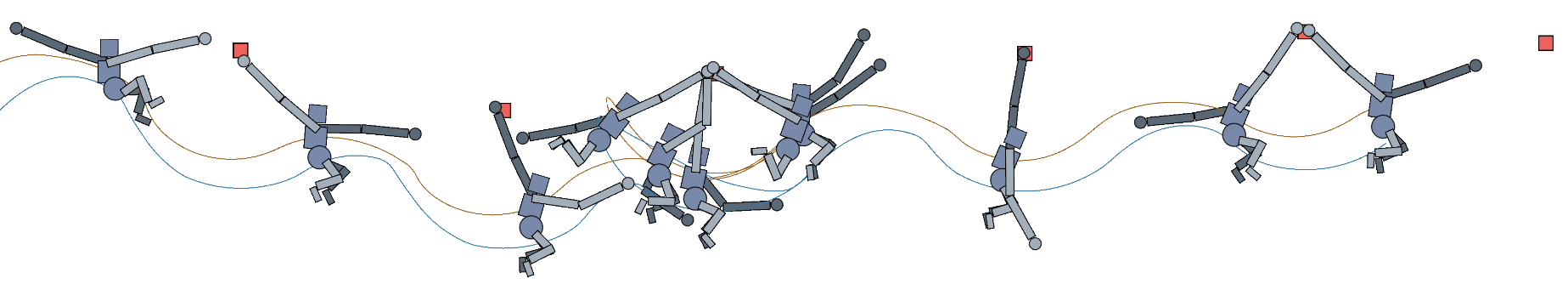}
  \includegraphics[width=.75\textwidth]{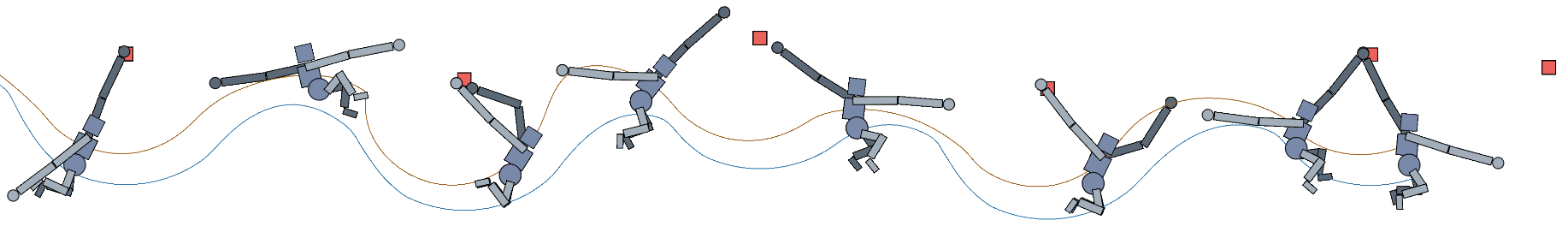}
  \includegraphics[width=.75\textwidth]{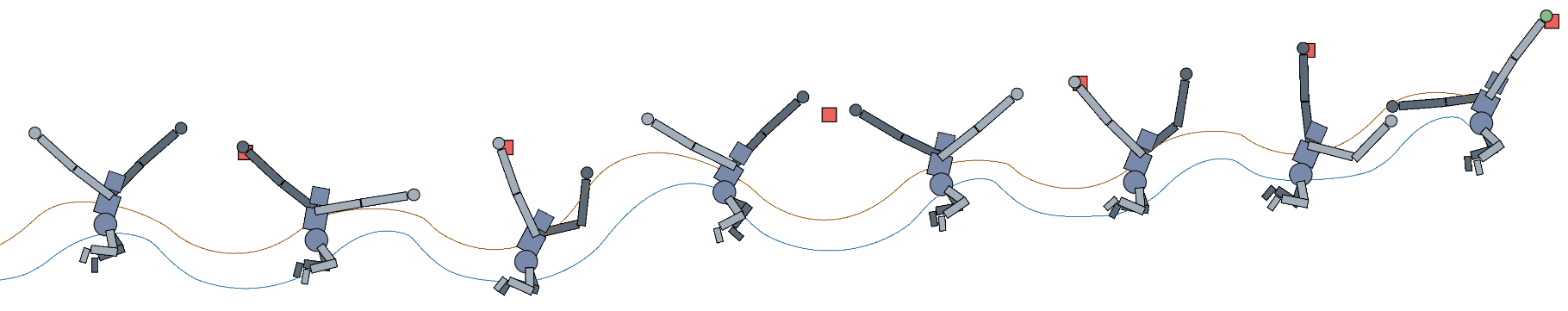}
  \includegraphics[width=.75\textwidth]{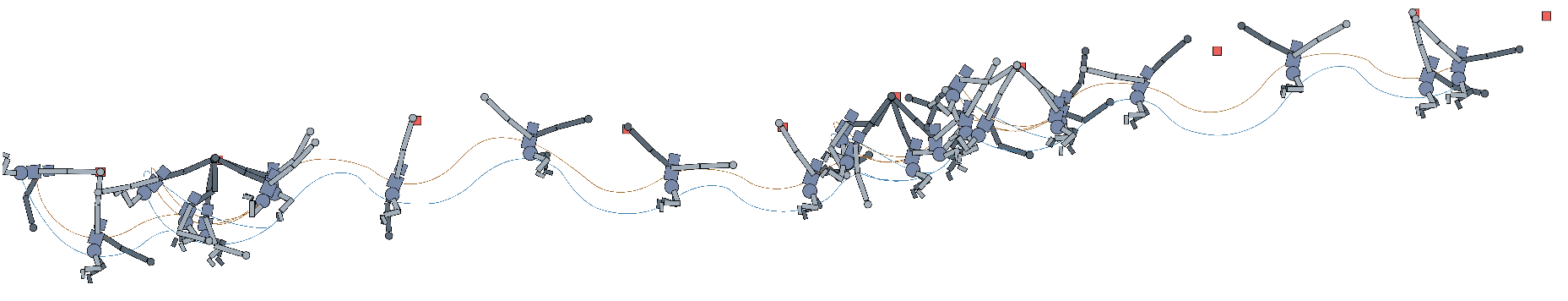}
  \caption{Additional trajectories showing a variety of motions.}
  \label{fig:more-motions}
\end{figure*}

\end{document}